\documentclass[lettersize,journal]{IEEEtran}
\usepackage{amsmath,amsfonts}
\usepackage{algorithmic}
\usepackage{algorithm}
\usepackage{array}
\usepackage[caption=false,font=normalsize,labelfont=sf,textfont=sf]{subfig}
\usepackage{textcomp}
\usepackage{stfloats}
\usepackage{url}
\usepackage{verbatim}
\usepackage{graphicx}
\usepackage{cite}
\usepackage{booktabs}
\usepackage{amssymb}
\usepackage{soul}
\usepackage{color}
\usepackage{pifont}
\usepackage{tikz}
\newtheorem{theorem}{Theorem}
\begin{document}

\title{

\vspace{-0cm} 
    \begin{tikzpicture}[remember picture, overlay]
        \node[anchor=north, yshift=-0.5cm] at (current page.north) {\fbox{\parbox{\textwidth}{\centering\small {\color{red}This work has been submitted to the IEEE for possible publication. Copyright may be transferred without notice, after which this version may no longer be accessible.}}}};
    \end{tikzpicture}
\vspace{0cm} 

Intrinsic Dynamics-Driven Generalizable Scene Representations for Vision-Oriented Decision-Making Applications}

\author{Dayang Liang, \textit{Graduate Student Member}, \textit{IEEE}, Jinyang Lai, \textit{Graduate Student Member}, \textit{IEEE}, \\and Yunlong Liu*, \textit{Member}, \textit{IEEE}
\thanks{This work was supported by the National Natural Science Foundation of China (No. 61772438 and No. 61375077).}
\thanks{Dayang Liang, Jinyang Lai, and Yunlong Liu are with the Department of Automation, School of Aerospace Engineering, Xiamen University, Xiamen 361005, China. (Corresponding author: Yunlong Liu)} 
\thanks{{\it Email address}: ylliu@xmu.edu.cn (Y. Liu)}
\thanks{ORCID (s): 0000-0001-8448-7845 (Y. Liu)}
\thanks{This work has been submitted to the IEEE for possible publication. Copyright may be transferred without notice, after which this version may no longer be accessible.}}

\markboth{}%
{Shell \MakeLowercase{\textit{et al.}}: A Sample Article Using IEEEtran.cls for IEEE Journals}


\maketitle

\begin{abstract}
How to improve the ability of scene representation is a key issue in vision-oriented decision-making applications, and current approaches usually learn task-relevant state representations within visual reinforcement learning to address this problem. While prior work typically introduces one-step behavioral similarity metrics with elements (e.g., rewards and actions) to extract task-relevant state information from observations, they often ignore the inherent dynamics relationships among the elements that are essential for learning accurate representations, which further impedes the discrimination of short-term similar task/behavior information in long-term dynamics transitions. To alleviate this problem, we propose an intrinsic dynamics-driven representation learning method with sequence models in visual reinforcement learning, namely DSR. Concretely, DSR optimizes the parameterized encoder by the state-transition dynamics of the underlying system, which prompts the latent encoding information to satisfy the state-transition process and then the state space and the noise space can be distinguished. In the implementation and to further improve the representation ability of DSR on encoding similar tasks, sequential elements’ frequency domain and multi-step prediction are adopted for sequentially modeling the inherent dynamics. Finally, experimental results show that DSR has achieved significant performance improvements in the visual Distracting DMControl control tasks, especially with an average of 78.9\% over the backbone baseline. Further results indicate that it also achieves the best performances in real-world autonomous driving applications on the CARLA simulator. Moreover, qualitative analysis results validate that our method possesses the superior ability to learn generalizable scene representations on visual tasks. The source code is available at https://github.com/DMU-XMU/DSR.

\end{abstract}


\section{Introduction}
\IEEEPARstart{V}{isual} reinforcement learning (DRL) has achieved great success in vision-oriented decision-making applications such as autonomous driving~\cite{10507015,10306329}, robot operations~\cite{9351748,10032100}, and video learning~\cite{10075396,gillberg2023technical} due to its ability to feature representation by the combination of neural networks~\cite{10426745}. However, a serious challenge faced by current DRL is that the policies trained in specific scenes struggle to generalize effectively to unknown scenes~\cite{wang2024investigating}. Moreover, it is exacerbated when policies are exposed to diverse scenarios. To tackle this challenge, previous work usually studies effective state representation learners to extract a representation space that can summarize task-relevant details within the environment, thus enabling the learning of generalizable policies~\cite{9779589}.

By generating diverse images to capture latent invariant information in anchor observations, data augmentation techniques have been widely employed in learning task-relevant representations and therefore discarding task-irrelevant components in the scene~\cite{10348509}. To be specific, noise contrast estimation (NCE) is often adopted as guidance to distinguish the task information and noise by optimizing different objectives \cite{10064325}, for example,  the Q-learning gradients with data augmentation for CG2A \cite{liu2023improving} and DrQ \cite{kostrikov2020image}, as well as the mutual information between pairs of positive and negative samples for TACO \cite{zheng2024texttt} and CURL \cite{laskin2020curl}. Additionally, some studies have proposed cyclic~\cite{ma2024learning} and domain augmentation~\cite{ryu2023instant} to enrich the spatial diversity in some scenarios.  However, these methods heavily rely on manually designed augmentation noises with weak universality. Moreover, it is impractical to design sufficient augmentation patterns to content with the generalization demands of real-world application scenes \cite{xia2022simgrace}.
 
Compared to data augmentation, Behavioral Similarity Metric (BSM) representation learning \cite{castro2020scalable} is more broadly applicable and requires no augmentation noise~\cite{10290946}. Deep bisimulation metric (DBC) \cite{zhang2020learning} is one of the prominent work in the literature, which optimizes a compact representation of the observation space by the distance of rewards and transition distributions. In practice, as such distances are difficult to optimize, conservative calculations \cite{liao2023policy} or relaxation \cite{chen2022learning} of the distances are often required.  Bisimulation-based clustering representation \cite{liu2023robust} or action-based similarity \cite{agarwal2020contrastive} is also adopted as a metric to be optimized for learning task-relevant representations. However, in these metric-based methods, the inherent dynamics relationships or constraints of the underlying system is ignored \cite{tang2023understanding}, resulting in distances to rewards, actions, or transition states being optimized independently. On the one hand, the learned representations may contain some task-irrelevant information as only one element used can not fully capture the true task in hand, on the other hand, due to the ignoring of the complete intrinsic dynamics constraints in the system, some crucial task-relevant information may be lost~\cite{liao2023policy}. Another limitation of the metric-based methods is that the using of only one-step predictions is not sufficient to effectively distinguish observations with similar short-term outcomes but different long-term transitions ~\cite{kemertas2021towards}.

\begin{figure*}[t]
\centering
\includegraphics[height=4.4cm]{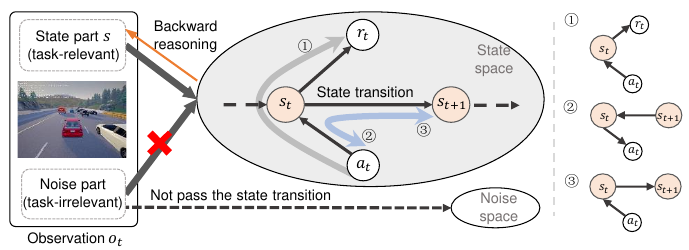}
\caption{Task-relevant state representation derived from dynamics relationships over underlying state transition in DRL.} 
\label{fig1}
\end{figure*}

To address these issues, driven by the intrinsic dynamics of the underlying system \cite{sutton2018reinforcement},  a representation method is proposed to separate the state space and noise space in image observations, where the intrinsic dynamics is used to constrain the optimization of the state encoder, which ensures that the encoded information progressively approximates the real state that obeys the state transition rule. We illustrate our motivation in Fig. \ref{fig1}. Specifically, in the DRL interactions, as actions are executed, the task-relevant true state parts in the underlying DRL system will transition from the current state to the next state according to a certain rule and thereby receive rewards, while task-irrelevant parts usually do not take part in this process\cite{sutton2018reinforcement}. Then, with the modeling of the intrinsic dynamics, i.e., the relationships among the states, actions, and rewards that in the state transition process, the latent task-relevant state information that adheres to the rule can be obtained. In details, we decompose the state transition into three parts: the reward model, the forward dynamic model, and the inverse dynamic model (as shown in the right part of Fig.~\ref{fig1}), and by optimizing the constrained objective in these models, the agent gradually focuses on and extracts task-relevant states.

In the actual implementations and to overcome the limitations of one-step metric predictions, discrete-time Fourier transform (DTFT) and multi-step latent prediction with global expressive capabilities \cite{ye2024state,10292587} are introduced to model the aforementioned three dynamics. In particular, for the reward and inverse dynamics, we utilize the state encoder to predict the Fourier frequency domain features of the low-dimensional action and reward sequence, while minimizing the distance between the prediction and the DTFT features of their corresponding true labels. This operation not only approximates the underlying system dynamics but also aids in extracting structured features from continuous observations. For the forward dynamics, we further introduce a latent overshooting (LO) \cite{hafner2019learning} model with the capability of long-term forward prediction to minimize the distance between the multi-step predicted state and the true state label, ensuring that the encoded information adheres to the law of forward dynamics. Moreover, this procedure, in conjunction with inverse dynamics, forms a cross-prediction and verification that enhances the accuracy of upstream encoded states.

DSR was initially evaluated across six tasks in the challenging Distracting DMControl Benchmark. It achieved significant overall performance improvements compared to the baselines, particularly with an average increase of 78.9\% over the DrQ baseline. Further, experiments demonstrated its superior performance in real-world autonomous driving applications in the CARLA environment, excelling in key metrics such as driving distance and collision. Lastly, qualitative analysis through t-SNE visualization confirmed the generalizable scene representation capacity of our method. In summary, extensive quantitative comparisons and qualitative analyses demonstrate that our method significantly enhanced its representational capabilities and policy performances in various vision-based decision-making applications.

Our contributions encompass three main manifolds:
\begin{itemize}
\item We propose an intrinsic dynamics-driven representation learning method that can exploit accurate task-relevant information constrained by full dynamics for generalization, which avoids the limitation of previous one-objective approaches that cannot extract complete task states. Moreover, different from previous data augmentation-related algorithms, no manual noise is required in our method.

\item We introduce the DTFT transform and latent overshooting model with global expressive capabilities to further handle long-term state prediction modeling of intrinsic dynamics, resulting in favorable structured feature extractions.

\item The proposed method achieves the best performances on both the Distracting DMControl Benchmark with video background distractions and the autonomous driving environment with natural visual distractions. Furthermore, the qualitative analysis results strongly demonstrate the outstanding scene representation abilities of the method.
\end{itemize}

\section{Preliminaries}
\subsection{Deep Reinforcement Learning}
In our task setting, the underlying interaction between the environment and the agent is modeled as a Markov decision process (MDP), described by the tuple ${\cal M}=({\cal S},{\cal A},{\cal P},R,\gamma)$, where $\cal S$ represents the continuous state space, $\cal A$ the continuous action space, ${\cal P}(s_{t+1}|s_t,a_t):{\cal S}\times {\cal A}\times {\cal S}\to[0,1]$ the probability of transitioning from state $s_t\in \cal{S}$ to state $s_{t+1}\in \cal{S}$ when executing an action $a_t\in \cal{A}$ at time $t$, $r_t = R(s_t,a_t):{\cal S}\times{\cal A}\to \mathbb{R}^1 \in {\cal R}$ the reward signal obtained by the agent when given the state $s_t$ and the action $a_t$, and $\gamma \in [0,1]$ the discount factor of rewards. 

An agent chooses an action $a_t\in {\cal A}$ according to a policy function $a_t~\pi\left(\cdot| s_t\right)$, which updates the system state $s_{t+1}\sim{\cal P}\left(\cdot| s_t,a_t\right)$, yielding a reward $r_t \in {\cal R}$. The agent’s optimization goal is to maximize the expected cumulative discounted rewards by learning a good policy: $\pi^{*} = {\underset{\pi}{\arg\max}{\mathbb{E}_{\tau\sim(\pi,{\cal M})}\left\lbrack {\sum_{t = 0}^{T - 1}{\gamma^{t}r_t}} \right\rbrack}}$. It is worth noting that due to the fact that single image observation usually does not satisfy the full observability of systems \cite{kaelbling1998planning}, we stack three consecutive images as the current observation $o_t$ while training an encoder network $\phi$ to extract true state information from the observation: $z_t=\phi(o_t)$.

\subsection{Actor-Critic Framework}

Actor-Critic (AC) framework \cite{konda1999actor} is an off-policy reinforcement learning algorithm for continuous control tasks, and it is also the fundamental framework for all algorithms mentioned in this work. In general, AC consists of a critic network for learning the value function $Q_\psi(s_t,a_t)$ and an actor network for learning the policy function $\pi_\varphi(a_t|s_t)$. Differ to the value-based methods, AC optimizes a non-deterministic policy to maximize the expected trajectory return.

Specifically, the critic network optimizes the temporal difference (TD) loss derived from the Bellman optimality equation while training the state action value function $Q_\psi(s_t,a_t)$ with parameter $\psi$ through sampling estimation \cite{mnih2015human}. The loss is defined as,
\begin{equation} 
\label{eq1}
{\cal L}{(\psi)}=\mathbb{E}_{s_t,a_t\sim \cal D}\left[\frac{1}{2}\left[(r_t+\gamma (1-d){\cal T}) - {\left(Q_{\psi}(s_t,a_t)\right)}^2\right]\right]
\end{equation}
with,
\begin{equation}
\label{eq2}
{\cal T}=\mathbb{E}_{a', s_{t+1}\sim \cal D}\left[{\underset{a'_t}\max}Q^{tgt}_{\hat{\psi}}(s_{t+1},a'_t)\right].
\end{equation}

In the above formula, $Q_{\hat{\psi}}^{tgt}$ represents the target $Q$ function with frozen network parameters $\hat{\psi}$, where $\hat{\psi}$ is updated from the exponential moving average (EMA) of the trainable parameters $\psi$, $d$ indicates the ``done'' signal of an episode, $\mathcal{D}$ represents the experience replay buffer of the off-policy DRL.

Furthermore, we train the actor-network with parameters $\varphi$ by sampling actions $a_t \sim\pi_\varphi$ from the policy and maximizing the expected reward of the sampled actions:
\begin{equation}
\label{eq3}
\mathcal{L}_{\pi}(\varphi,\psi) = \mathbb{E}_{a^{'}\sim\pi_{\varphi},s_{t}\sim\mathcal{D}}\left\lbrack {Q^{\pi}_\psi\left( s_{t},a' \right)} \right\rbrack
\end{equation}

\subsection{Behavior Similarity Metrics}
Behavior Similarity Metric (BSM) uses elements such as rewards and actions to measure task similarity between two states, with a typical representative of Bisimulation Metric \cite{zhang2020learning}. Bisimulation Metric measures the equivalence relationship between two states in a recursive form, i.e., if two states share an equivalence distribution on the next equivalent state, as well as they have the same immediate reward, they are considered equivalent \cite{larsen1989bisimulation}, and then a task-relevant state encoder can be learned with such a Bisimulation Metric~\cite{zang2024understanding}. 

Bisimulation Metric-based ideas are adopted in our approach, but we further utilize the dynamics of the underlying system to refine the task-relevant representation, which is more task-focused~\cite{liao2023policy} and can be easily extended to the multi-step prediction cases to alleviate the limitations of the currently used one-step metric.

\subsection{Discrete-Time Fourier Transform}
The Discrete-Time Fourier Transform (DTFT) is a powerful continuous signal processing tool that converts discrete-time signals into continuous frequency domain information. Formally, the DTFT converts a discrete real sequence ${\{x_n\}}_{n=-\infty}^{+\infty}$ into a frequency domain signal through a complex variable function $F(\omega)=\sum_{n=-\infty}^{\infty}{x_ne^{-(j\omega n)}}$, where $\omega$ is a frequency domain variable, $j$ is an imaginary unit, and $F\left(\omega\right)$ satisfies the periodic property, i.e., $F\left(\omega\right)=F\left(\omega+2k\pi\right)$. 

In this work, we leverage this DTFT to reveal the intrinsic dynamics relationship among continuous time-domain signals (e.g., sequential acitons) from a frequency domain perspective, thereby facilitating the encoder to capture global features such as periodicity and structure. To simplify the calculation, we focus on the frequency domain signal of $\omega$ within a single period interval $[-\pi,\pi]$ that still contains all information in the infinite time domain.

\section{Method}

In this section, we propose an intrinsic dynamics-driven representation learning method with sequence models, namely DSR, for improving the performance in generalizable scene representation and policy in visual reinforcement learning. Specifically, we first introduce the intrinsic dynamics regarding motivation and give the corresponding sequential optimization objective in DSR. Finally, We further introduce its internal frequency domain prediction and multi-step forward dynamics for sequential modeling. Note that since DSR only uses the inherent properties of RL itself to learn efficient feature representations, it can be extended to DRL frameworks that are not limited to purely visual inputs, and is applicable to arbitrary decision-making applications.

\begin{figure*}[bh]
\centering
\includegraphics[height=4.6cm]{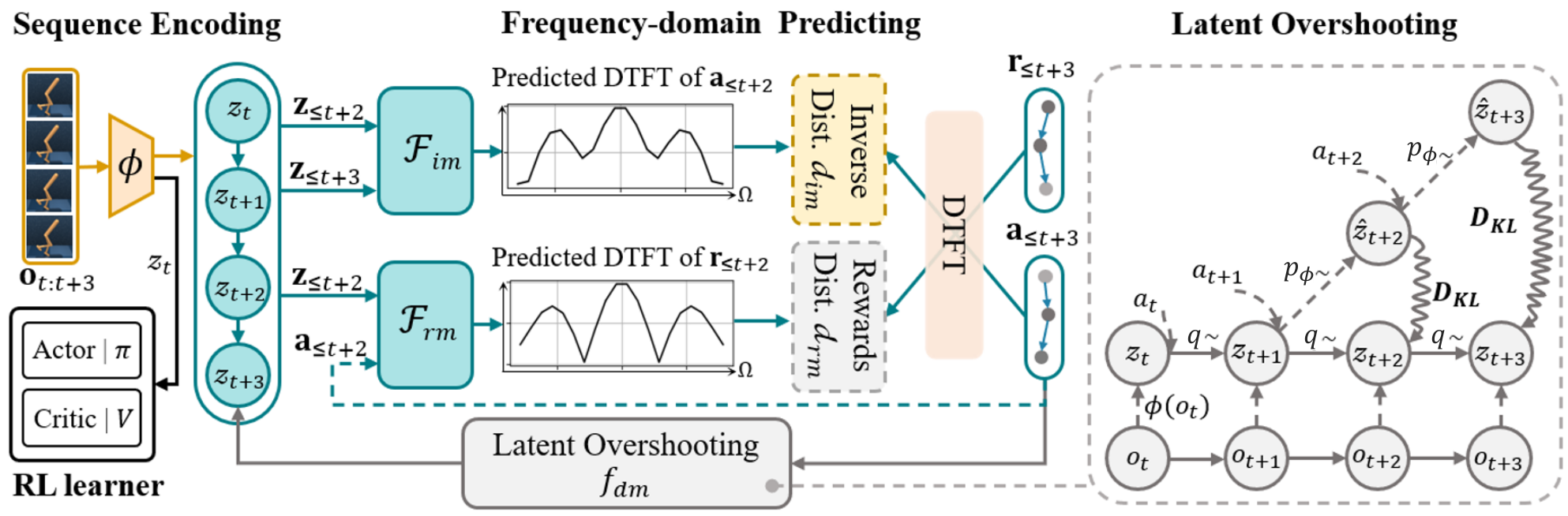}
\caption{Overview of the DSR framework: The method is divided into four parts: sequence encoding, frequency domain prediction, latent overshooting, and reinforcement learning, with different sections distinguished by colored connection arrows. The entire framework focuses on the encoder $\phi$ as the core training target, and the trained encoder will be used for reinforcement learning policy training. In the figure, $z_{\le t+2}$ and $z_{\le t+3}$ are shorthand for the sequential latent encoded states  $z_{t:t+2}$ and $z_{t+1:t+3}$, respectively (similarly for other vectors).}
\label{fig2}
\end{figure*}

\subsection{Intrinsic Dynamics}
The intrinsic dynamics of the underlying reinforcement learning system can be defined as a probability distribution function $T\left(s_{t+1},r_t\middle|s_t,a_t\right):{\cal S}\times{\cal A}\times{\cal S}\times{\cal R}\rightarrow[0,1]$[22], which can be decomposed into \ding{172}reward, \ding{173}inverse and \ding{174}forward dynamics as shown in the right part of Fig. \ref{fig1} and further describe as Eq. (\ref{eq4}):

\begin{equation}
\label{eq4}
\left\{ \begin{matrix}
\left. {}r_{t} = r\left( {s_{t},a_{t}} \right),r:{\cal S} \times {\cal A}\rightarrow {\cal R} \right. \\
\left. {}a_{t} = v\left( s_{t},s_{t + 1} \right),v:{\cal S} \times {\cal S}\rightarrow {\cal A} \right. \\
\left. {}s_{t + 1} = p\left( s_{t},a_{t} \right),p:{\cal S} \times {\cal A}\rightarrow {\cal S} \right.
\end{matrix} \right.
\end{equation}

Functions $r$, $v$, and $p$ are the real dynamics of the underlying system. As mentioned, these three dynamics are related to true states and they can be modeled to extract the latent encoding state (task-relevant representation) from observations: $z_t=\phi(o_t)$, where $\phi$ is the encoder that should be learned. Accordingly, the optimized objectives can be further defined as follows:

\begin{equation}
\label{eq5}
\left\{ \begin{matrix}
\left. {}r_{t} \simeq r_{\phi}\left( {z_{t},a_{t}} \right),~r_{\phi}:Z \times {\cal A}\rightarrow {\cal R} \right. \\
\left. {}a_{t} \simeq v_{\phi}\left( z_{t},z_{t + 1} \right),v_{\phi}:Z \times Z\rightarrow {\cal A} \right. \\
\left. {}z_{t + 1} \simeq p_{\phi}\left( z_{t},a_{t} \right),p_{\phi}:Z \times {\cal A}\rightarrow Z \right.
\end{matrix} \right.
\end{equation}
where the reward $r_\phi$, reverse dynamics $v_\phi$, and forward dynamics $p_\phi$ are respectively modeled as prediction models with reward $r_t$, action $a_t$, and encoding state $z_{t+1}$ as self-supervised labels. 

Then, these models with dynamics relationships will be integrated into the following one-step learning objective ${\cal J}^{one}(\phi)$,
\begin{align} 
\label{eq6}\nonumber
&{\cal J}^{one}(\phi)= d\left( {r_{\phi}\left( {z_{t},a_{t}} \right),{\overset{-}{r}}_{t}} \right) + d\left( {v_{\phi}\left( {z_{t},z_{t + 1}} \right),{\overset{-}{a}}_{t}} \right)  \\
&+ d\left( {p_{\phi}\left( {z_{t},a_{t}} \right),{\overset{-}{z}}_{t + 1}} \right)
\end{align}
where $d$ represents specific optimization distance defined later, and the self-supervised label ${\bar{z}}_{t+1}$ is a frozen encoding state, which uses bootstrapping to improve the encoder parameters.

\subsection{Sequential Optimization in DSR}
While Eq. (\ref{eq6}) can be directly optimized to learn the task-relevant representation $z_t=\phi(o_t)$, its one-step distance cannot effectively distinguish scenarios with similar short-term outputs but different long-term behaviors, especially in environments with sparse rewards~\cite{kemertas2021towards}. To address this issue, with the benefits of easily extending to multi-step predictions for DSR, the multi-step optimization loss objective is employed as shown in Eq. (\ref{eq7}): 
\begin{align} \label{eq7}
\nonumber
&{\cal J}^{seq}(\phi)= d\left( {r_{\phi}\left( \mathbf{z}_{t:t + T - 1},\mathbf{a}_{t:t + T - 1} \right),\mathbf{r}_{t + 1:t + T}} \right) \\\nonumber
&+ d\left( {v_{\phi}\left( {\mathbf{z}_{t:t + T - 1},\mathbf{z}_{t + 1:t + T}} \right),\mathbf{a}_{t + 1:t + T}} \right) \\
&+ d\left( {p_{\phi}\left( \mathbf{z}_{t:t + T - 1},\mathbf{a}_{t:t + T - 1} \right),{\overset{-}{z}}_{t + T}} \right)
\end{align}

Moreover, as favorable periodic and highly structured timing-related features usually exist in sequential inputs~\cite{ye2024state}, the adoption of the sequential optimization objective allows the encoder to learn task-related temporal features.

Specifically, we introduce the Discrete Time Fourier Transform (DTFT) to obtain the frequency domain signals of action sequences $\mathbf{a}_{t+1:t+T}\in\mathbb{R}^{T\times d_1}$ and reward sequences $\mathbf{r}_{t+1:t+T}\in\mathbb{R}^{T\times1}$ in the T dimension, where $d_1$ and $T$ are the action dimension and the element sequences' length respectively. In addition, for high-dimensional ${\bar{z}}_{t+T}\in\mathbb{R}^{d_2}$ ($d_2$ is the dimension of the encoding state), we utilize a latent overshooting model that satisfies the forward dynamics to minimize the Kullback-Leibler (KL) divergence between the multi-step state prediction and the real state distribution. Then, we get the final loss to be minimized for optimization:
{\small 
\begin{align}
\label{eq8} \nonumber
&\mathcal{L}(\phi) = d_{rm}\left( {\mathcal{F}_{rm}^{\phi}\left( {\mathbf{z}_{t:t + T - 1},\mathbf{a}_{t:t + T - 1}} \right),F\left( \mathbf{r}_{t + 1:t + T} \right)} \right) \\\nonumber
&+ d_{im}\left( {\mathcal{F}_{im}^{\phi}\left( {\mathbf{z}_{t:t + T - 1},\mathbf{z}_{t + 1:t + T}} \right),F\left( \mathbf{a}_{t + 1:t + T} \right)} \right) \\
&+ D_{KL}\left\lbrack p_\phi\left(z_{t+T}\middle|\mathbf{o}_{t:t+T-1},\mathbf{a}_{t:t+T-1}\right) \middle| |q\left(z_{t+T}\middle|\mathbf{o}_{t:t+T-1},\mathbf{a}_{t:t+T-1}\right) \rbrack \right.
\end{align}
}where $\mathcal{F}_{rm}^\phi$ and $\mathcal{F}_{im}^\phi$ represent the Fourier prediction models for the reward and inverse dynamics respectively, $F$ denotes the Fourier transform of the sequence target, $p_\phi\left(z_{t+T}\middle|\mathbf{o}_{t:t+T-1},\mathbf{a}_{t:t+T-1}\right)$ represents the probability distribution of the predicted multi-step states, $q\left(z_{t+T}\middle|\mathbf{o}_{t:t+T-1},\mathbf{a}_{t:t+T-1}\right)$ represents the true state distribution probability, ``$d$'' represents the Mean Square Error (MSE) distance, and subscripts ``$rm$'' and ``$im$'' represent the reward model and the inverse dynamic model. 

The whole framework of our approach is shown in Fig.~\ref{fig2}. For details of the frequency domain prediction model and the multi-step state prediction, we discuss them in the following two subsections.

\subsection{Prediction Models for Frequency Domain}

The inverse model with sequential actions is used to illustrate the details of its self-supervised DTFT target and its Fourier features' prediction process, the same procedure is employed for the reward model case.

First, since the frequency domain period of the discrete-time action signal is $2\pi$, we only consider the DTFT over a single period $[-\pi, \pi]$. The following equation represents the complex-valued function $F_\mathbf{a}\left(\mathbf{\omega}\right)$ for the frequency variable $\mathbf{\omega}$ in the DTFT of the sequence action target:
\begin{equation} 
\label{eq9} 
F_{\mathbf{a}}\left( \mathbf{\omega} \right) = {\sum_{n = t + 1}^{t + T}{a_{n}e^{- in\mathbf{\omega}}, }}\mathbf{\omega} \in \lbrack - \pi, \pi\rbrack
\end{equation} 
where $\mathbf{\omega}\in\mathbb{R}^k$ is a list of frequencies sampled evenly $k$ times over the interval  $[-\pi,\pi]$, where $k=20$. 

In practice, the amplitude $\Gamma_{Amp}F_\mathbf{a}\in\mathbb{R}^{1\times k}$ and phase $\Gamma_{Pha}F_\mathbf{a}\in\mathbb{R}^{1\times k}$ of the action's DTFT are used as self-supervised targets, defined as follows:
\begin{equation} 
\label{eq10} 
\Gamma_{Amp}F_{\mathbf{a}} = \left| {F_{\mathbf{a}}(\omega)} \right|,~\Gamma_{Pha}F_{\mathbf{a}} = arctan\frac{Im\left( F_{\mathbf{a}}(\omega) \right)}{Re\left( F_{\mathbf{a}}(\omega) \right)}
\end{equation}

For the Fourier prediction model, based on the settings of the inverse dynamics Eq. (\ref{eq5}), we use the adjacent sequence encodings $(\mathbf{z}_{t:t+T-1},\mathbf{z}_{t+1:t+T})$ as an input to the inverse dynamics parametric model $f_\phi\in\mathbb{R}^{2\times k}$, denoted by $\mathcal{F}_{im}$ as follows:
\begin{equation} \small
\label{eq11} 
\mathcal{F}_{im}\left( {\mathbf{z}_{t:t + T - 1},\mathbf{z}_{t + 1:t + T}} \right) = f_{\phi}\left( {agg\left( {\phi\left( \mathbf{o}_{t:t + T - 1} \right),\phi\left( \mathbf{o}_{t + 1:t + T} \right)} \right)} \right)
\end{equation} 

Ultimately, by integrating the aforementioned DTFT target $F_\mathbf{a}$ and the prediction model $\mathcal{F}_{im}$, we optimize both the amplitude distance and phase distances of the sequential actions to ensure the inverse dynamics process:
{\small\begin{align} 
\label{eq12} 
\nonumber
&d_{im}\left( {\mathcal{F}_{im}\left( {\mathbf{z}_{t:t + T - 1},\mathbf{z}_{t + 1:t + T}} \right),F\left( \mathbf{a}_{t + 1:t + T} \right)} \right) \\
&= \mathbb{E}_{a\sim\pi,s\sim {\cal P}}\left\lbrack {\left\| {\Gamma_{Amp}\mathcal{F}_{im} - \Gamma_{Amp}F_{\mathbf{a}}} \right\| + \left\| {\Gamma_{Pha}\mathcal{F}_{im} - \Gamma_{Pha}F_{\mathbf{a}}} \right\|} \right\rbrack
\end{align} }

Similarly, we can derive the optimization loss for the reward prediction model, which targets the DTFT features of the sequential rewards $E_{\binom{1}{2}}$:
{
\small\begin{align} 
\label{eq13} 
\nonumber
&d_{rm}\left( {\mathcal{F}_{rm}\left( \mathbf{z}_{t:t + T - 1},\mathbf{a}_{t:t + T - 1} \right),F\left( \mathbf{r}_{t + 1:t + T} \right)} \right) \\
&= \mathbb{E}_{a\sim\pi,s\sim {\cal P}\atop r\sim {\cal R}}\left\lbrack {\left\| {\Gamma_{Amp}\mathcal{F}_{rm} - \Gamma_{Amp}F_{\mathbf{r}}} \right\| + \left\| {\Gamma_{Pha}\mathcal{F}_{rm} - \Gamma_{Pha}F_{\mathbf{r}}} \right\|} \right\rbrack
\end{align} }

\begin{algorithm}[t]
    \caption{DSR}
    \label{algorithm1}
    \textbf{Input}: replay buffer ${\cal D}$ with size $N$, learning rate, etc. \\
    \textbf{Output}: optimal $\pi$
    \begin{algorithmic}[1] 
    		\STATE Initialize Critic network $\varphi$, Actor network $\psi$, and  encoder network $\phi$.
		\FOR{episode $m \gets 0$ to $M$}
        	\STATE {Initial observation $o_t$.}
			\STATE Encode state: $z_t=\phi(o_t)$.
			\STATE Cumpute action: $ a_t=\pi(\cdot|z_t)$.
			\STATE Excute an interaction: $ o_t,a_t,r_{t+1},done = Env(a_t)$.
			\STATE Collect data ${\cal D}\gets {\cal D}\cup \{o_t,a_t,r_{t+1},done\}$. 
		\ENDFOR
		\FOR{gradient step $i \gets 0$ to $I$}
			\STATE Get sequence $\{a_{i:i+T-1},r_{i:i+T-1},o_{i:i+T-1}\}$ from $D$.
			\STATE Transform action sequence by DTFT in Eq. (\ref{eq11}).
			\STATE Transform reward sequence by DTFT in Eq. (\ref{eq11}).
			\STATE Compute the objectives of $d_{im}$, $d_{rm}$ and $f_{dm}$.
			\STATE Train the Actor-Critic ${\cal L}_{Q}+{\cal L}_{\pi}$. (Eq. (\ref{eq1}) and Eq. (\ref{eq3})).
			\STATE Train the auxiliary task ${\cal L}=d_{im}+d_{rm}+f_{dm}$. (Eq. (\ref{eq8})).
		\ENDFOR
        \STATE \textbf{return} optimal $\pi$
    \end{algorithmic}
\end{algorithm}

\subsection{Forward dynamics}
To learn a latent encoding state that complies with the forward dynamics, we introduce a long-term forward dynamic model based on multi-step planning, known as Latent Overshooting \cite{hafner2019learning}, as shown on the right side of Fig. \ref{fig2}. Overall, LO uses a two-level planner to better understand the dynamic transition process through long-term behaviors to build a state transition model with long-term predictive accuracy, encouraging the encoder to extract task-relevant information in the forward dynamics.

The above optimization problem can be characterized as minimizing the KL divergence between the multi-step latent state transition distribution $p_\phi\left(z_{t+T}\middle|\mathbf{o}_{t:t+T-1},\mathbf{a}_{t:t+T-1}\right)$ and the true state distribution $q\left(z_{t+T}\middle|\mathbf{o}_{t:t+T-1},\mathbf{a}_{t:t+T-1}\right)$, denoted as $D_{KL}\left[p_\phi||q\right]$. However, as the original KL distance includes the intractable true distribution $q$, we use variational inference to transform this challenging problem into a parameterized gradient optimization problem. For this, we first present the following theory:
\begin{theorem}
Let the sequence observation data be represented by $\mathbf{o}_{t:t+T-1}$ and  $\mathbf{a}_{t:t+T-1}$, and given the approximate posterior $p_\phi$ with the encoding parameters $\phi$, the evidence lower bound (ELBO) on the data log-likelihood is:
{\small
\begin{align}
\label{eq14} 
\nonumber
&logq\left( {\mathbf{o}_{t:t + T - 1},\mathbf{a}_{t:t + T - 1}} \right) \\\nonumber
&= {\int{p_{\phi}\left( z_{t + T} \middle| {\mathbf{o}_{t:t + T - 1},\mathbf{a}_{t:t + T - 1}} \right)logq\left( {\mathbf{o}_{t:t + T - 1},\mathbf{a}_{t:t + T - 1}} \right)dz_{t + T}}} \\\nonumber
&= \mathbb{E}_{z_{t + T}\sim p_{\phi}}log\frac{q\left( {z_{t + T},\mathbf{o}_{t:t + T - 1},\mathbf{a}_{t:t + T - 1}} \right)}{p_{\phi}\left( z_{t + T} \middle| {\mathbf{o}_{t:t + T - 1},\mathbf{a}_{t:t + T - 1}} \right)} \\\nonumber
&+ \mathbb{E}_{z_{t + T}\sim p_{\phi}}D_{KL}\left\lbrack p_{\phi} \middle| \middle| q \right\rbrack \\\nonumber
&\geq \mathbb{E}_{z_{t + T}\sim p_{\phi}}log\frac{q\left( {z_{t + T},\mathbf{o}_{t:t + T - 1},\mathbf{a}_{t:t + T - 1}} \right)}{p_{\phi}\left( z_{t + T} \middle| {\mathbf{o}_{t:t + T - 1},\mathbf{a}_{t:t + T - 1}} \right)} \\
&\triangleq ELBO\left( {\mathbf{o}_{t:t + T - 1},\mathbf{a}_{t:t + T - 1}} \right)
\end{align}}
\end{theorem}

\begin{IEEEproof}
The proof can be seen in Appendix A2.
\end{IEEEproof}

Based on this conclusion, since the evidence probability $logq\left(\mathbf{o}_{t:t+T-1},\mathbf{a}_{t:t+T-1}\right)$ remains constant during the optimization of the posterior $p_\phi$, minimizing the original objective $D_{KL}\left[p_\phi||q\right]$ is equivalent to maximizing the $ELBO\left(\mathbf{o}_{t:t+T-1},\mathbf{a}_{t:t+T-1}\right)$. Hence, we can further derive the objective $f_{dm}$ as follows (``$dm$'' represents forward dynamics model):

{\small
\begin{align}
\label{eq15}
\nonumber
&\min f_{dm}={{D_{KL}\left\lbrack p_{\phi} \middle| \middle| q \right\rbrack}} = {{ELBO\left( \mathbf{o}_{t:t + T - 1},\mathbf{a}_{t:t + T - 1} \right)}} \\\nonumber
&= \mathbb{E}_{z_{t + T}\sim p_{\phi}}log\frac{q_{\theta}\left( {\mathbf{o}_{t:t + T - 1},\mathbf{a}_{t:t + T - 1},z_{t + T}} \right)}{p_{\phi}\left( z_{t + T} \middle| {\mathbf{o}_{t:t + T - 1},\mathbf{a}_{t:t + T - 1}} \right)} \\\nonumber
&= \mathbb{E}_{z_{t + T}\sim p_{\phi}}log\frac{q_{\theta}\left( {\mathbf{o}_{t:t + T - 1},\mathbf{a}_{t:t + T - 1}} \right)q_{\theta}\left( z_{t + T} \middle| \mathbf{o}_{t:t + T - 1},\mathbf{a}_{t:t + T - 1} \right)}{p_{\phi}\left( z_{t + T} \middle| {\mathbf{o}_{t:t + T - 1},\mathbf{a}_{t:t + T - 1}} \right)} \\\nonumber
&= \mathbb{E}_{z_{t + T}\sim p_{\phi}} \underset{reconstruction}{\underbrace{logq_{\theta}\left( {\mathbf{o}_{t:t + T - 1},\mathbf{a}_{t:t + T - 1}} \right)}} \\
&- D_{KL} \underset{multi - step~prediction}{\underbrace{\left. p_{\phi}\left( z_{t + T} \middle| {\mathbf{o}_{t:t + T - 1},\mathbf{a}_{t:t + T - 1}} \right) \middle| \middle| q_{\theta}\left( z_{t + T} \middle| \mathbf{o}_{t:t + T - 1},\mathbf{a}_{t:t + T - 1} \right) \right.}} 
\end{align}
}

Ultimately, the original challenge is transformed into a gradient-optimizable problem, involving the KL loss between the priors associated with encoding parameters $\phi$ and reconstruction parameters $\theta$, and a reconstruction loss term related to $\theta$.

It is important to note that in the early stages of training, using the imprecise latent state ${\bar{z}}_{t+T}$ as the self-supervised target for the forward dynamics model $p_\phi\left(z_{t+T}\middle|\mathbf{o}_{t:t+T-1},\mathbf{a}_{t:t+T-1}\right)$ will lead to instability in representation learning. To this end, we incorporate an adaptive factor $\delta_{\pi,t}$ into the above objective, thus obtaining the final objective:
{
\begin{align}
\nonumber
&\min f_{dm}={{D_{KL}^{\delta}\left\lbrack p_{\phi} \middle| \middle| q \right\rbrack}} = {{ELBO^{\delta}\left( \mathbf{o}_{t:t + T - 1},\mathbf{a}_{t:t + T - 1} \right)}} \\\nonumber
&= \mathbb{E}_{z_{t + T}\sim p_{\phi}}{\left\lbrack {logq_{\theta}\left( {\mathbf{o}_{t:t + T - 1},\mathbf{a}_{t:t + T - 1}} \right)} \right\rbrack} - \delta_{\pi,t}D_{KL}{\left\lbrack p_{\phi} \middle| \middle| q_{\theta} \right\rbrack}
\end{align}
}

The aim is to reduce the early influence of the forward dynamics item on overall representation learning. $\delta_{\pi,t}$ is calculated as follows:
\begin{equation} 
\label{eq16} 
\delta_{\pi,t} = min\left( {\rho_{\pi,t},clip\left( {\rho_{\pi,t},1 - \epsilon,1 + \epsilon} \right)} \right)
\end{equation} 
with,
\begin{equation} 
\label{eq17} 
\rho_{\pi,t} = \frac{1}{\left| \mathcal{A} \right|}{\sum_{i = 1}^{|\mathcal{A}|}\left| \frac{c}{\pi_{\theta + \mathrm{\Delta}\theta}^{(i)}\left( z_{t} \right) - \pi_{\theta}^{(i)}\left( z_{t} \right)} \right|}
\end{equation} 

To obtain the factor $\delta_{\pi,t}$ at time $t$, we first need to calculate the average difference $\rho_{\pi,t}$ between the current policy $\pi_{\theta + \mathrm{\Delta}\theta}^{(i)}\left( z_{t} \right)$ and the old policy $\pi_\theta^{\left(i\right)}\left(z_t\right)$ under the same current encoding state $z_t$, where $c$ is a scaling hyperparameter. Additionally, to keep it within a reasonable positive range, we use the clip function to trim $\rho_{\pi,t}$. Overall, the contribution of the forward dynamics item to the overall representation learning increases as the policy improves.

\section{Experiments}

To test the scene representation ability and policy performance of the DSR method, we first conduct standardized evaluations on six continuous control tasks on the challenging vision-based Distracting DeepMind Control Suite benchmark (DMControl) \cite{stone2021distracting}. Further, to verify the superiority of DSR, we applied it to the real-world autonomous driving application within the CARLA-v0.98 simulation environment \cite{dosovitskiy2017carla}. In addition, we also qualitatively discussed the generalization ability of DSR to task-relevant information in visual tasks by the results of t-SNE visualization. In the whole experiment, DSR is compared with a series of recent representation-related DRL baselines that perform strongly on vision-based control tasks.  It is worth noting that the DSR can be embedded into a variety of visual DRL frameworks as an efficient plug-in. In this work, DSR is extended to the DrQ framework \cite{kostrikov2020image}, so it is also marked as DrQ+DSR. One focus of the experiments is to observe the performance improvement produced by inserting the DSR method into the DrQ.

\textbf{Baselines} The DrQ-based DSR method takes advantage of the inherent dynamics to learn representations. Therefore, we first compared the recent DrQ algorithm \cite{kostrikov2020image}, which achieves SOTA performances on the clean DMControl benchmark through a regularized Q function. As a comparison regarding dynamics representation, we compare the PAD algorithm \cite{hansen2020self}, which achieves superior generalization performances using a forward dynamics model. Finally, we compared the common CURL \cite{laskin2020curl}, which primordially introduced contrastive representation learning into visual DRL, while achieving the best performances on the DMControl benchmark.

\subsection{Evaluation on the Distracting DMControl suite}

\begin{figure}[h]
\centering
\includegraphics[height=2.1cm]{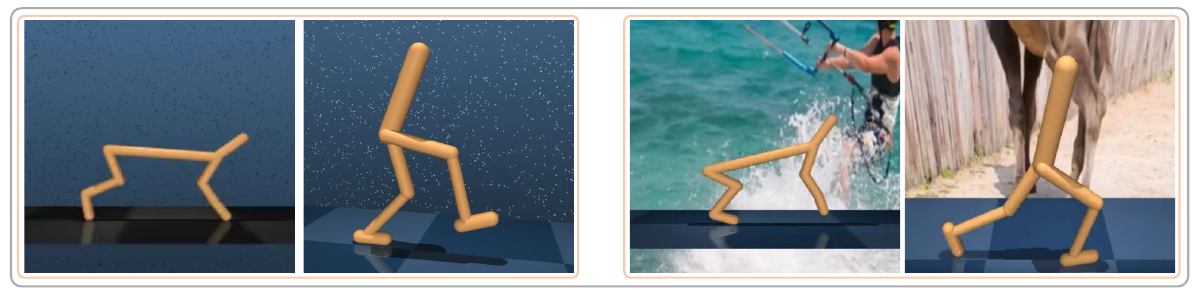}
\caption{Visual task examples for cheetah\_run and walker\_walk in DMControl. \textbf{Left}: clean DMControl setting with original background; \textbf{Right}: distracting DMControl setting with random background videos. Among them, the cheetah\_run task is to control the six-degree-of-freedom cheetah robot to run rapidly.; the walker\_walk task is to control the six-degree-of-freedom humanoid robot to walk rapidly.}
\label{fig4}
\end{figure} 

\textbf{Distracting DeepMind Control Suite Setting} Distracting DMControl suite is a hard setting with video background distracting grounded in the DeepMind Control Suite continuous control task \cite{tassa2018deepmind}, which has been widely used in visual DRL to test the performance of scene generalization. Following the settings of existing work \cite{stone2021distracting}\cite{yang2022learning}, the distracting DMControl suite replaces the clean background with natural video sampled from the DAVIS 2017 dataset \cite{pont20172017} as the noise of the observations. Furthermore, to verify the generalization ability of the method in unseen scenes, it is trained in two fixed distracting videos, but evaluated on arbitrary scenes among 30 unseen distracting videos.

\begin{figure}[b]
\centering
\includegraphics[height=5.45cm]{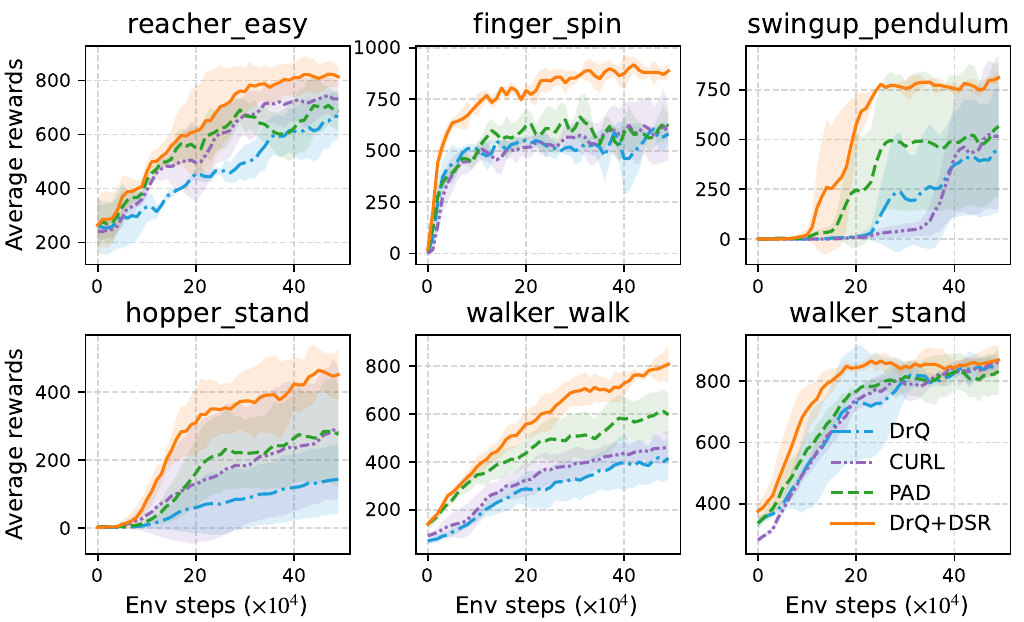}
\caption{Evaluation curves on distracting DMControl suite with unseen video background setting at 500K environment steps. For each method, the results are derived from the mean rewards and standard deviation of 3 random seed experiments. A total of 50 evaluations were completed for each experiment, where the checkpoint score for each evaluation was averaged over 10 episodes. The yellow line (DrQ+DSR) is our method.}
\label{fig3}
\end{figure}

\textbf{Hyperparameter Setting} We use the same hyperparameter settings and the network architecture for DSR and baselines. Specifically, the training observations are stacked by three sequential RGB frames with the shape $9 \times 84 \times 84$. For the individual DSR representation task, we set the sequence length of MDP elements to 3, the sample number of the DTFT period to 20, and the minibatch of the auxiliary training to 256. Detailed parameters are shown in Table \ref{tab:4} of Appendix A1.

\begin{table*}[th]\normalsize
    \centering
    \caption{The best mean episode rewards on distracting DMControl suite with unseen video background setting at 500K environment steps. We report the mean and standard deviation of three repeated experiments, as well as the improvement percentage of DSR compared to the backbone DrQ.)}
    \begin{tabular}{lcccccc}
        \toprule
        Methods      & Reacher\_easy & Finger\_spin & Swingup\_pendulum & Hopper\_stand & Walker\_walk & Walker\_stand \\
        \midrule
        DrQ          & 668.5$\pm$86 & 606.3$\pm$7 & 452.4$\pm$320 & 143.8$\pm$103 & 418.0$\pm$93 & 847.9$\pm$22 \\
        CURL         & 745.0$\pm$68 & 623.3$\pm$119 & 524.6$\pm$370 & 289.3$\pm$207 & 461.2$\pm$73 & 863.0$\pm$20 \\
        PAD          & 707.7$\pm$69 & 664.0$\pm$67 & 567.1$\pm$352 & 284.7$\pm$152 & 610.8$\pm$92 & 832.9$\pm$77 \\ \midrule
        \textbf{DSR (ours)}   & \textbf{824.0$\pm$46} & \textbf{917.9$\pm$138} & \textbf{811.1$\pm$33} & \textbf{464.7$\pm$63} & \textbf{808.4$\pm$79} & \textbf{869.3$\pm$47} \\
        \textbf{vs. DrQ}          & $\left. \uparrow 23.3\mathbf{\%} \right.$ & $\left. \uparrow 51.4\mathbf{\%} \right.$ & $\left. \uparrow 79.3\mathbf{\%} \right.$ & $\left. \uparrow 223.2\mathbf{\%} \right.$ & $\left. \uparrow 93.4\mathbf{\%} \right.$ & $\left. \uparrow 2.5\mathbf{\%} \right.$ \\
        \bottomrule
    \end{tabular}
    \label{tab1}
\end{table*}
\begin{figure*}[th]
\centering
\includegraphics[height=3.4cm]{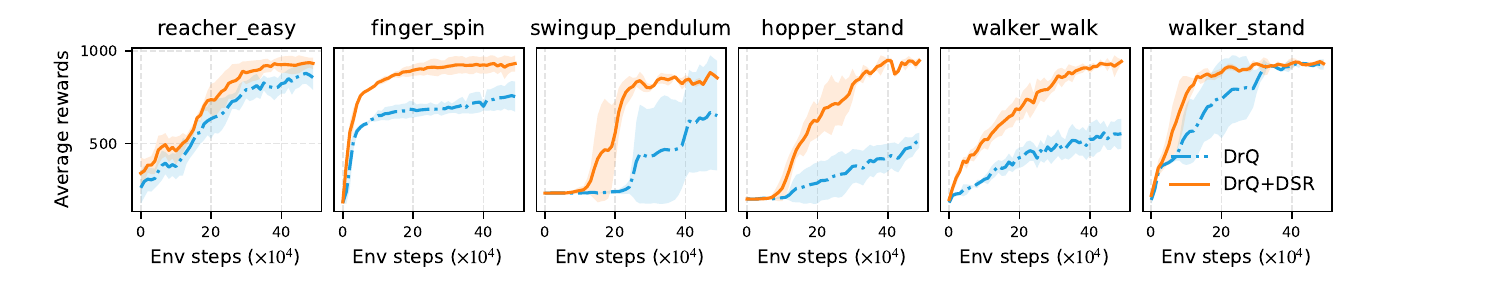}
\caption{Training curves of our method (DrQ+DSR) and DrQ backbone on two seen background videos.}
\label{fig5}
\end{figure*} 
\begin{figure*}[bh]
\centering
\includegraphics[height=5.0cm]{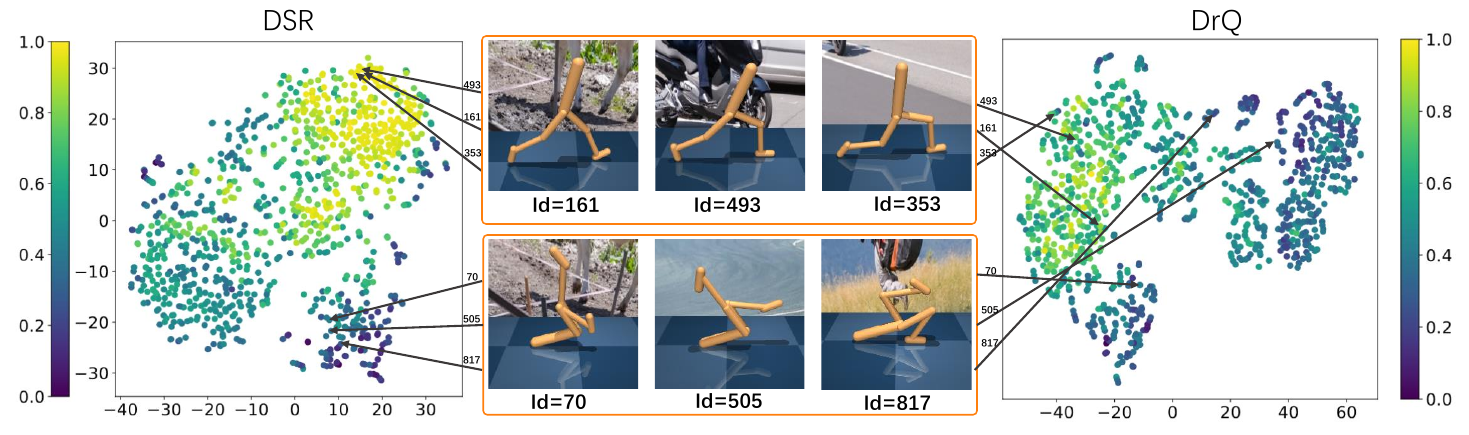}
\caption{t-SNE visualizations of the latent encoding spaces learned with the DSR encoder
(left t-SNE result) and DrQ encoder (right t-SNE result). Each observation image points to positions (higher value yellow, lower value purple) under the two encoding spaces through the corresponding arrow.
}
\label{fig6}
\end{figure*}

\textbf{Main Results}
As depicted in Fig. \ref{fig3} and  Table \ref{tab1}, we report both the evaluation curves and the best results of DSR compared to the recent DrQ, CURL, and PAD on 6 tasks of reacher\_easy, finger\_spin, swingup\_pendulum, hopper\_stand, walker\_walk and waker\_stand over the process of training. As a result, the proposed DSR method achieves significant performance improvements in all tasks compared to all baselines, especially compared to the DrQ backbone, which averaged 78.9\%. In addition to the best rewards, DSR also converged faster in comparison to all baselines. Intuitively, the comparative results powerfully demonstrate that DSR has learned task-relevant state information during the training process, so that it is able to execute background-independent policies in unseen scenarios, and then achieve the best policy performance. Notably, the swingup\_pendulum and hopper\_stand are sparse reward tasks with few dense reward signals, which wreaks havoc on both policy and representation learning. Nevertheless, the DSR method still improves 79.3\% and 223.2\% compared to the DrQ, as shown in Table \ref{tab1}.

As mentioned above, evaluation results show that our method improves DrQ's abilities in both task-relevant representation and generalization to unknown scenes/videos, yet the abilities need to be learned and acquired during training. To further demonstrate the learning ability of the DSR, we provide a separate comparison between DSR and DrQ in the training process in Fig. \ref{fig5}. We can clearly observe that once DSR is inserted into DrQ, the proposed DrQ+DSR method greatly improves the overall performance of the original DrQ in terms of convergence speed and optimal rewards on 6 visual tasks.

\textbf{Qualitative Analysis}
To qualitatively analyze the representation ability of the encoder, we employ t-SNE visualization to illustrate that DSR can encode observation images with similar behaviors/tasks into close coordinate distances, achieving the extraction of invariant representations independent of background distractions. In this experiment, we use t-SNE to project the encoding state of batch observations under the DSR or DrQ encoder space into a two-dimensional space and then embed them into Cartesian coordinates. In addition, we also use the learned state value to render the color of each coordinate point. Therefore, the position of each coordinate point represents the encoded information of an observation, and the color depth represents the value of the observation.
 
As shown in Fig. \ref{fig6}, the two sides are the t-SNE embedding plots of the same batch of observation images under the DSR and DrQ encoders respectively, and the middle is two arbitrary observation groups with similar tasks in the batch. From the figure, we can easily see that, i) for each observation group, although the background noise between observations is very different, their coordinate distances (left) in the DSR encoding space are very close. However, they are far apart in the DrQ encoding space (right). This result intuitively shows that it is precisely because the encoder trained by DSR can extract task-relevant information more accurately than DrQ that it can encode the observation group with similar tasks into a very close distance. ii) By observing the color of the embedding position, we also find that similar tasks/behaviors encoded by DSR have similar values, showing consistently high values for good behaviors (upper group) and low values for unfavorable behaviors (lower group). Instead, this is not entirely consistent in the DrQ embedding plot. The comparative result shows that our method can learn a more accurate state value function that is beneficial to policy learning.

\subsection{Verification on the Autonomous Driving}
 \begin{figure}[h]
\centering
\includegraphics[height=4.0cm]{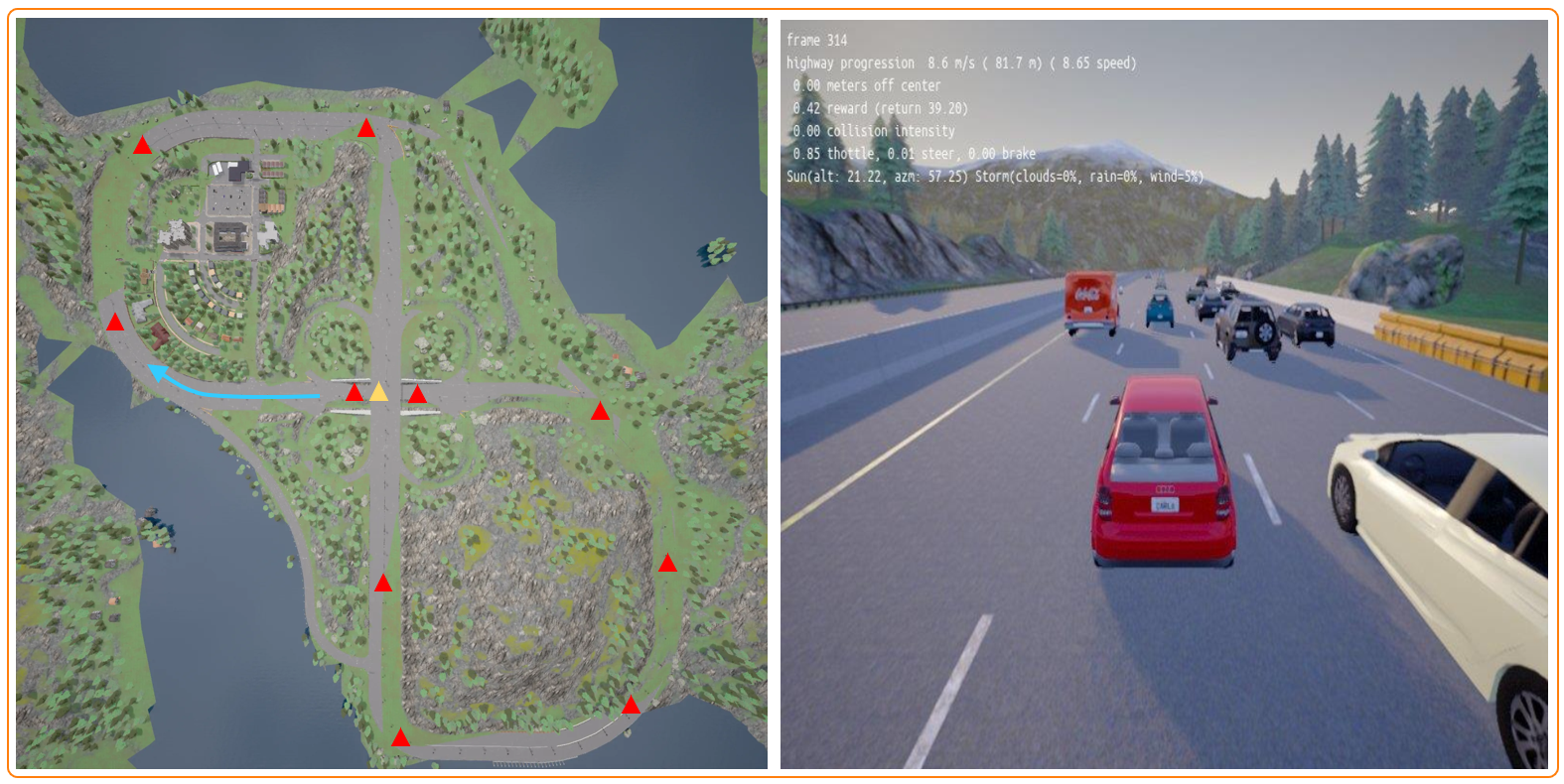}
\caption{\textbf{Left}: An aerial view of the CARLA-Town04 map, in which triangles represent some of the officially provided spawn/start points, and the blue arrow represents the driving direction; \textbf{Right}: A live photo, in which the red vehicle in the center is the trained vehicle agent and the others are NPC vehicles.}
\label{fig7}
\end{figure}

To further verify the scene representation ability and policy performance of DSR in natural scenes, we test the experiment on CARLA, a hyper-realistic autonomous driving simulator environment that simulates various real components and natural phenomena. According to existing work, the experiments are implemented on the Town04 map in the CARLA simulator, depicted in Fig. \ref{fig7}. Our goal is to use first-person visual images as observations to learn a driving policy that can control the vehicle to travel the longest distance. Since the driving scene contains a large number of task-irrelevant elements (e.g. mountains, clouds, and rain), as well as task-relevant feature information (e.g. roads, vehicles, and obstacles), this requires the agent to be able to learn an invariant generalization representation space.

Compared with existing work \cite{zhang2020learning,chen2022learning}, our task setting is more difficult. Specifically, the previous methods reset to a fixed starting position for each training episode, i.e., the yellow triangle position in the aerial view (Fig. \ref{fig7} left), which greatly limits the effective driving scenarios. In our setting, the starting points are randomly sampled from about 100 spawn points (triangular positions in Fig. \ref{fig7}) distributed throughout the map, which can effectively test the agent's generalization ability to diverse scenarios. Another difference is that the training observation of DSR is only composed of three cameras on the vehicle’s roof, which is a 180° wide-angle RGB image with a size of $3\times84\times252$ pixels. Other settings are consistent with existing work. For example, the reward is set to $\left. r_{t}\left( {s, a} \right) = v^{T}\hat{\mathbf{u}}*\mathrm{\Delta}t - C_{i}*impulse - C_{s}* \middle| steer \right|$ represents the effective speed of the vehicle projected onto the highway, impulse is the impact force ($N/s$) obtained by sensors, and $|steer|$ is the amplitude of each direction control.

\begin{figure}[h]
\centering
\includegraphics[height=2.9cm]{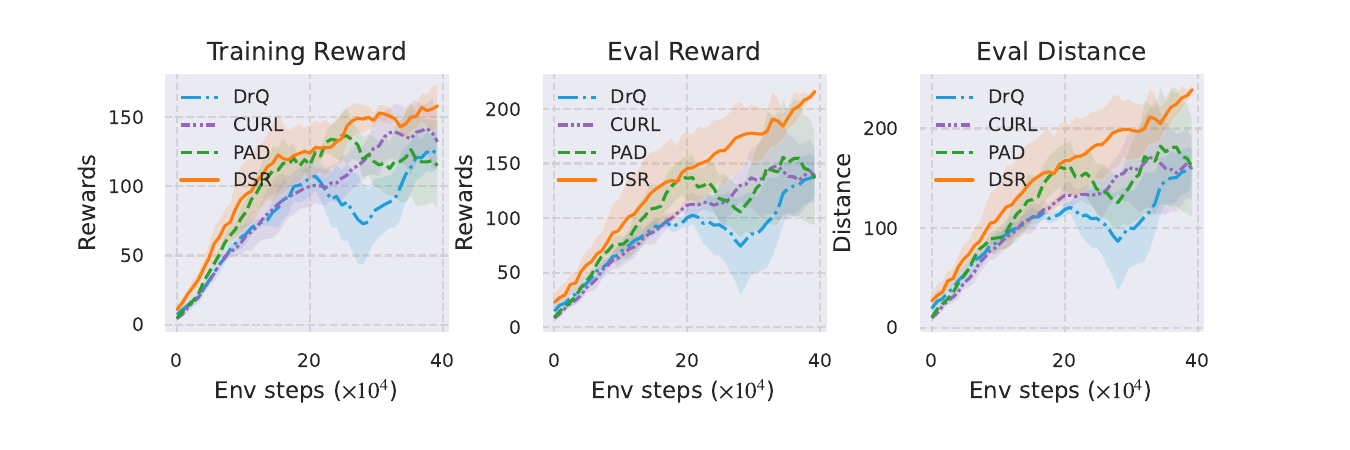}
\caption{Experimental results of methods in the CARLA driving environment at 400K environment steps. We report the average and standard deviation of each method under three random seeds.}
\label{fig8}
\end{figure}

\textbf{Main Results.}
Fig. \ref{fig8} and Table \ref{tab2} show the main comparison results on driving control tasks. Specifically, in Fig. \ref{fig8}, we report the learning curves for the average episode reward (left), the evaluation curves corresponding to the average episode reward evaluated at every 10k step interval (center), and the evaluation curves to the average episode distance (right). Also, we first report the best scores on key driving metrics (i.e., Distance and Reward items) in Table \ref{tab2}. The results show that, although the starting points are expanded to the entire Town04 map with various scenes, our DSR method still achieves the best reward and distance in both the training and the evaluation process. Especially during the evaluation process, the driving distance of DSR is 31.2\% higher than the best baseline with a clear advantage.

\begin{figure}[h]
\centering
\includegraphics[height=2.45cm]{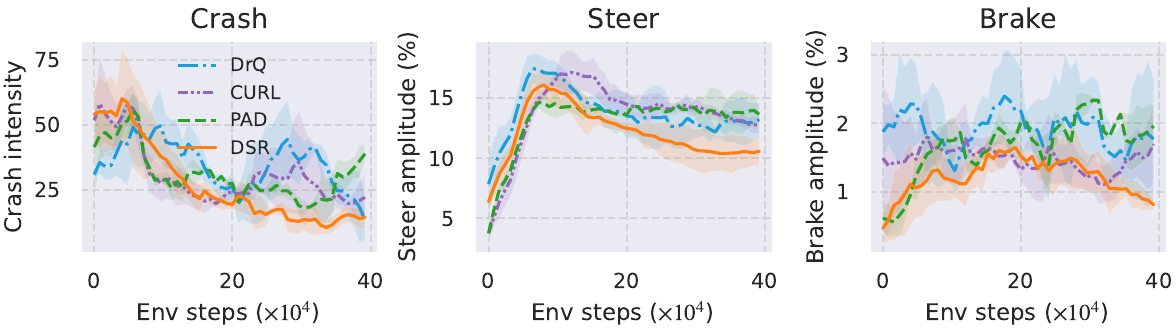}
\caption{Comparison of key performance metrics over the training process in the CARLA driving environment.}
\label{fig9}
\end{figure} 

\begin{table*}[th]\normalsize
    \centering
    \caption{Comparison of best episode scores in the CARLA driving environment. We recorded the highest scores for the first three items and the lowest scores for the last three items (at convergence).}
    \begin{tabular}{lcccccc}
        \toprule
        Methods      & Distance & Eval-Reward & Train-Reward & Mean Crash (N/s) & Mean Steer(\%) & Mean Brake (\%) \\
        \midrule
        DrQ          & 161.3$\pm$18 & 138.8$\pm$18 & 128.1$\pm$15 & 13.5$\pm$8 & 12.2$\pm$0.5 & 1.5$\pm$0.4 \\
        CURL         & 170.9$\pm$59 & 147.5$\pm$50 & 141.8$\pm$18 & 19.8$\pm$1 & 12.7$\pm$0.7 & 1.1$\pm$0.3 \\
        PAD          & 182.3$\pm$59 & 155.7$\pm$53 & 136.4$\pm$18 & 18.1$\pm$3 & 13.4$\pm$1.6 & 1.4$\pm$0.5 \\
        \textbf{DSR (ours)}   & \textbf{239.1$\pm$5} & \textbf{216.0$\pm$6} & \textbf{158.0$\pm$15} & \textbf{10.6$\pm$3} & \textbf{10.4$\pm$1.4} & \textbf{0.8$\pm$0.1} \\
        \textbf{vs. best scores}   & \textbf{$\left. \uparrow 31.2\mathbf{\%} \right.$} & \textbf{$\left. \uparrow 38.7\mathbf{\%} \right.$} & \textbf{$\left. \uparrow 11.4\mathbf{\%} \right.$} &  &  &  \\
        \bottomrule
    \end{tabular}

    \label{tab2}
\end{table*}

\begin{figure*}[th]
\centering
\includegraphics[height=6.7cm]{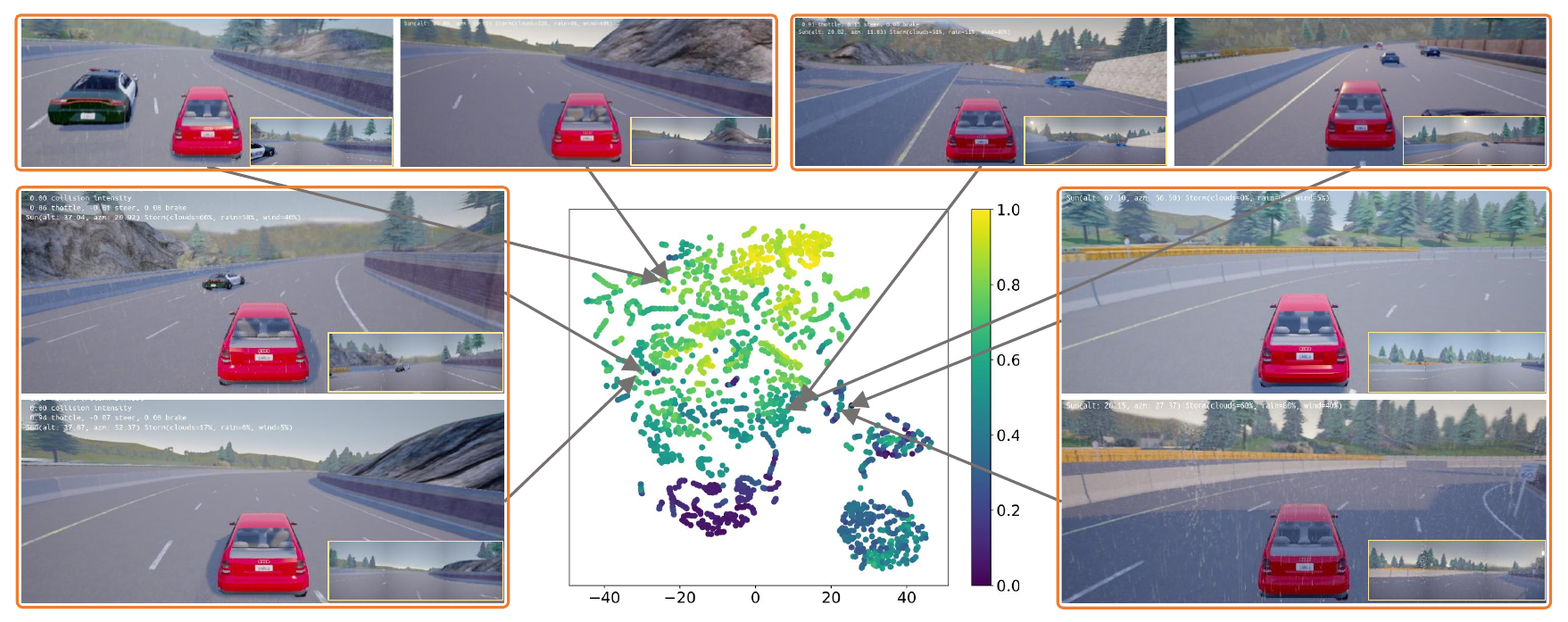}
\caption{ t-SNE visualization of the latent encoding space learned with our DSR encoder in the CARLA driving task. The red vehicles in the live photos are the trained driving agents, and the insets on the lower right are the training observations synthesized by the first-person perspective cameras.}
\label{fig10}
\end{figure*}

In addition, we recorded the data changes of the collision, steering, and braking performance metrics related to driving agents with the policy learning process, as shown in Table \ref{tab2} and Fig. \ref{fig9}. It is worth noting that our reward function is set with the goal of encouraging the agents to drive longer distances. However, all methods collectively learned to reasonably control the magnitude of the steering wheel and brake operation while reducing collisions. More importantly, as shown in Fig. \ref{fig9} and Table \ref{tab2}, the proposed method achieved the smallest average collision force, and the most reasonable direction and braking control scheme after the policy converges. In Table \ref{tab2}, "Steer" and "Brake" respectively represent the percentage of the direction angle amplitude and the percentage of the braking amplitude controlled by each action step.

\textbf{Qualitative Analysis.} To verify the ability of DSR to encode task-relevant feature information in the real-world driving scenes, we still used t-SNE to visualize the encoding distance of a batch of high-dimensional image observations under the DSR encoding space, as shown in Fig. \ref{fig10}. The center of Fig. \ref{fig10} is the two-dimensional embedding plot of a batch of observations in the encoding space, and the periphery is four groups of observations with similar tasks/behaviors but large differences in natural backgrounds.

It is easy to observe that for the observation groups with very similar task information (e.g., lower right: dangerous sharp right turn; upper left: off-right turn), our DSR encoder can still encode each group to a very close distance, even though the natural backgrounds of each group are very different (e.g., lower right: bright sunny day vs. dim rainy day; upper left: trees and other vehicles vs. mountains). The result strongly demonstrates that the DSR encoder can still accurately extract task-relevant information even in complex natural backgrounds. Furthermore, this capability can promote the generalization for natural scenes and even solve some unseen risk corner cases in visual autonomous driving.

\section{Related Work}

\textbf{Visual Reinforcement Learning.}~~An important challenge in visual reinforcement learning is how to learn generalizable representations based on image inputs in vision-based control applications \cite{10041739,10145773}. Previous work has made significant efforts in this area, such as early focuses on end-to-end Attention \cite{josef2020deep} representation learning. Their basic principle relies on the reinforcement learning loss to train a task-relevant feature weight map, thereby extracting key features from the original image \cite{liang2021gated}. However, end-to-end representations that share a single reinforcement learning loss are prone to gradient vanishing issues during backpropagation, leading to insufficient representation learning ability \cite{laskin2020curl}. Recently, many studies have focused on auxiliary representations, such as constructing an independent self-supervised loss through contrastive learning \cite{9939161}, encoding reconstruction \cite{10105983}, attention mechanisms\cite{10219179}, or behavior similarity metric \cite{zhang2020learning}. Their advantage is that more complex and effective representation learning tasks can be designed without interfering with the gradient propagation of reinforcement learning itself \cite{laskin2020curl}. In this work, we construct an auxiliary representation learning loss by intrinsic dynamics relationships to learn an efficient encoder. 

\vspace{9pt}
\textbf{Data Augmentation Representations.}~~Data augmentation typically involves operations such as rotation, random cropping, random masking, and random PaResize on observational images \cite{ma2024learning}, enriching data diversity while combining specific methods to achieve effective feature representations \cite{10173582,zhang2020learning}. For example, in previous work, contrastive representation methods combined with data augmentation use InfoNCE loss to maximize the mutual information between an anchor and its augmented positive samples, while minimizing the mutual information between the anchor and negative samples to learn invariant representations of the augmented samples \cite{9701666}. However, most contrastive representation methods are limited to using random negative samples, which lacks a mechanism to select negative samples that contain task-relevant information, thus limiting their representation ability in complex visual scenes \cite{liang2023sequential}. Additionally, data augmentation is often used to improve their evaluation variance by regularizing Q-functions \cite{kostrikov2020image}, eliminate task-irrelevant gradient information by balancing the Q-learning gradients of augmented samples \cite{liu2023improving}, and achieve an understanding of dynamics features by pixel-level or vector-level reconstruction \cite{10105983}. Nevertheless, most data augmentation work relies on predefined image augmentation operations, while we cannot design infinite augmentation paradigms for real-world scenarios. Moreover, since the augmentation operations in some scenarios do not fully satisfy information invariance, it is hard to achieve effective representation learning. In contrast, our method utilizes state transition relationships for representation learning, which can avoid the aforementioned issues.

\vspace{9pt}
\textbf{Metric-based Representations.}~~Behavior Similarity Metric (BSM) methods are based on bisimulation theory \cite{ferns2004metrics}, which employ elements such as actions on MDP collected from DRL interactions, to measure and extract equivalent task information from noisy observations \cite{larsen1989bisimulation}. Because these methods leverage the intrinsic properties of reinforcement learning to learn representations without relying on additional knowledge, they have been widely studied recently \cite{10476686}. DBC is an earlier BSM method that uses reward and state transition distances to extract task-relevant information \cite{zhang2020learning}, but its strict distance measures can over-optimize the encoder, potentially losing beneficial information \cite{liao2023policy}. To address this, Chen et al. proposed a relaxed bisimulation metric, i.e., RAP distance \cite{chen2022learning}, which significantly improved optimization efficiency based on bisimulation theory. Additionally, some studies suggest using actions \cite{liang2023sequential}, rewards \cite{ye2024state}, and other interpretable combinations \cite{yuan2024learning} to extract equivalent task-relevant information. However, most of these methods are limited to one-step distance, which makes they difficult to infer different task information at similar element distances \cite{kemertas2021towards}. To address these issues, we introduced sequence modeling methods to strongly identify similar one-step task/behavior information. Additionally, to improve the optimization of our objective, we use a self-supervised prediction mode instead of differential distance \cite{liang2023sequential} of elements. 

\vspace{9pt}
\textbf{Multi-step Representations.}~~Although multi-step prediction representations can encode single-step task information more accurately, learning beneficial features through long-term prediction is challenging due to the difficulty of learning an accurate long-term prediction model \cite{ye2024state}. To address this, further work models characteristic functions of rewards or action distributions in high-dimensional spaces \cite{yang2022learning,liang2023sequential} to learn potential global structured features. Additionally, some work uses frequency domain features of reward or discounted state transitions \cite{ye2024state} as auxiliary objectives to accelerate the extraction of potential temporal representations. In model-based deep reinforcement learning, Hafner et al. introduced Latent Overshooting, a multi-step state prediction technique that optimizes a sequential VAE \cite{zhu2020s3vae} to enhance the model's understanding of forward dynamics \cite{hafner2019learning}. However, multi-step prediction rarely focuses on the interrelationships between different elements (e.g., the relationships between action and reward on MDP), instead calculating distances derived from each element independently, making it difficult for the encoder to leverage constraints between elements to extract more accurate states. Therefore, our method not only leverages the underlying dynamics transition relationships among different elements but also introduces useful sequence modeling methods to constrain encoder training.

\section{Conclusion and Future Work}

We proposed an efficient state representation learning method within the DrQ DRL framework that sequentially models the dynamics relationships associated with the underlying state transition process for improving the ability of scene generalization and policy performance in vision-oriented decision-making applications. The method leverages the underlying state transition rules of the system, i.e., driven by the action, the state information existing in the observation spontaneously transitions to the next state and receives the corresponding reward, whereas irrelevant non-state information typically does not satisfy this transition, thereby efficiently decomposing task-relevant state information and noise from observations. 

DSR was first evaluated on the challenging Distracting DMControl control tasks, showing the best policy performance and especially an average improvement of 78.9\% over DrQ baseline, while also achieving the best control performance in real-world autonomous driving applications with visual natural noise. Additionally, qualitative analysis results based on t-SNE visualization confirmed that the DSR method has more efficient scene representation abilities compared to baselines. In summary, DSR can effectively extract key features in visual scenes and further enhance policy performance, and thus it can be used as an effective plug-in within DRL to promote potential research in vision-oriented decision-making applications.

Although our method has achieved significant improvements in scene representation and policy performance, it consumed additional computational resources to handle multi-step serial planning in the forward dynamics model. To address this limitation, in the future, we will further investigate how to improve the computational efficiency of the forward dynamics model while ensuring model accuracy. In addition, we are also further verifying the control performance of our method in real-world autonomous driving.


{\appendices

\section*{A1. Parameters on Distracting DMControl and CARLA tasks}

\begin{table}[H]\normalsize\small
    \centering
    \caption{Key hyperparameters for experiments of DMControl tasks}\label{tab:4}
    \begin{tabular}{ll}
        \toprule
        Hyperparameters for DMControl tasks         & Value  \\
        \midrule
        Observation downsampling           & 84 $\times$ 84           \\
        Training frames      & 500000           \\
        Replay buffer capacity   & 100000           \\
       Stacked frames     & 3           \\
         Batch size         & 256           \\
          Learning rate  & 0.0005           \\
          Discount factor & 0.99 \\	 
         Init temperature         & 0.1           \\
		Scaling hyperparameter & 0.6 \\
          Element sequence length  & 3           \\
          State representation dimension & 50 \\
         Optimizer         & Adam           \\
         \midrule
         Special hyperparameters for Driving tasks        &   \\
         \midrule
        Camera number                            & 3  \\
        Full fov angles                               &  3$ \times $60 degree   \\
        Observation downsampling           & 84 $\times$ 252           \\
        Training frames      & 400000           \\
         Batch size         & 64           \\
          $\Delta t$             &   0.05 seconds          \\
         $C_i$              &  0.0001       \\
         $C_s$             &   1.0           \\
        \bottomrule
    \end{tabular}
\end{table}

\section*{A2. Proof of Evidence Lower Bound on Sequential Data}
\begin{IEEEproof}
First, we introduce the parameterized posterior probability distribution $
p_{\phi}\left( z_{t + T} \middle| {\mathbf{o}_{t:t + T - 1},\mathbf{a}_{t:t + T - 1}} \right)$, and subsequently derive the log-likelihood probability $logq\left( \mathbf{o}_{t + n - 1},\mathbf{a}_{t + n - 1} \right)$ of the data as:
\begin{align}\nonumber
 \label{eq19} 
&logq\left( {\mathbf{o}_{t:t + T - 1},\mathbf{a}_{t:t + T - 1}} \right) \\\nonumber
&= {\int{p_{\phi}\left( z_{t + T} \middle| {\mathbf{o}_{t:t + T - 1},\mathbf{a}_{t:t + T - 1}} \right)logq\left( {\mathbf{o}_{t:t + T - 1},\mathbf{a}_{t:t + T - 1}} \right)dz_{t + T}}} \\\nonumber
&= \mathbb{E}_{z_{t + T}\sim p_{\phi}}logq\left( {\mathbf{o}_{t:t + T - 1},\mathbf{a}_{t:t + T - 1}} \right) \\\nonumber
&= \mathbb{E}_{z_{t + T}\sim p_{\phi}}logq\left( {z_{t + T},\mathbf{o}_{t:t + T - 1},\mathbf{a}_{t:t + T - 1}} \right) \\\nonumber
&- \mathbb{E}_{z_{t + T}\sim p_{\phi}}logq\left( z_{t + T} \middle| {\mathbf{o}_{t:t + T - 1},\mathbf{a}_{t:t + T - 1}} \right) \\\nonumber
&= \underset{ELBO{({\mathbf{o}_{t:t + T - 1},\mathbf{a}_{t:t + T - 1}})}}{\underbrace{\mathbb{E}_{z_{t + T}\sim p_{\phi}}log\frac{q\left( {z_{t + T},\mathbf{o}_{t:t + T - 1},\mathbf{a}_{t:t + T - 1}} \right)}{p_{\phi}\left( z_{t + T} \middle| {\mathbf{o}_{t:t + T - 1},\mathbf{a}_{t:t + T - 1}} \right)}}} \\
&+ \underset{D_{KL}{\lbrack{p_{\phi}||q}\rbrack}}{\underbrace{\mathbb{E}_{z_{t + T}\sim p_{\phi}}log\frac{p_{\phi}\left( z_{t + T} \middle| {\mathbf{o}_{t:t + T - 1},\mathbf{a}_{t:t + T - 1}} \right)}{q\left( z_{t + T} \middle| {\mathbf{o}_{t:t + T - 1},\mathbf{a}_{t:t + T - 1}} \right)}}}
\end{align}

In the above Eq. (\ref{eq19}), the log probability of evidence is decomposed into the sum of the evidence lower bound and the KL distance of the posterior. Since the evidence probability $logq\left( {\mathbf{o}_{t:t + T - 1},\mathbf{a}_{t:t + T - 1}} \right)$ does not contain the posterior information $
z_{t + T}$, it is fixed when optimizing the parameterized posterior $
p_{\phi}\left( z_{t + T} \middle| {\mathbf{o}_{t:t + T - 1},\mathbf{a}_{t:t + T - 1}} \right)$. Therefore, considering that the latter term always holds $D_{KL}\left\lbrack p_{\phi} \middle| \middle| q \right\rbrack \geq 0$, we then obtain the following:
{
\begin{align}
 \label{eq20} 
\nonumber
&logq\left( {\mathbf{o}_{t:t + T - 1},\mathbf{a}_{t:t + T - 1}} \right) \\\nonumber
&= \mathbb{E}_{z_{t + T}\sim p_{\phi}}log\frac{q\left( {z_{t + T},\mathbf{o}_{t:t + T - 1},\mathbf{a}_{t:t + T - 1}} \right)}{p_{\phi}\left( z_{t + T} \middle| {\mathbf{o}_{t:t + T - 1},\mathbf{a}_{t:t + T - 1}} \right)} + D_{KL}\left\lbrack p_{\phi} \middle| \middle| q \right\rbrack \\\nonumber
&\geq \mathbb{E}_{z_{t + T}\sim p_{\phi}}log\frac{q\left( {z_{t + T},\mathbf{o}_{t:t + T - 1},\mathbf{a}_{t:t + T - 1}} \right)}{p_{\phi}\left( z_{t + T} \middle| {\mathbf{o}_{t:t + T - 1},\mathbf{a}_{t:t + T - 1}} \right)} \\
&\triangleq ELBO\left( {\mathbf{o}_{t:t + T - 1},\mathbf{a}_{t:t + T - 1}} \right)
\end{align}
}

\end{IEEEproof}

}
\bibliographystyle{IEEEtran}
\bibliography{ieee}


\vfill

\end{document}